\documentclass{article}

\usepackage{microtype}
\usepackage{graphicx}
\usepackage{subfigure}
\usepackage{booktabs} 
\usepackage{calrsfs}
\usepackage{amsmath}
\usepackage{multirow}

\usepackage{amssymb}
\usepackage{mathtools}
\DeclareMathAlphabet{\pazocal}{OMS}{zplm}{m}{n}

\DeclarePairedDelimiter\floor{\lfloor}{\rfloor}

 \usepackage{balance}
 \usepackage{flushend}

\usepackage{hyperref}

\usepackage[accepted]{icml2019}

\begin{document}

\twocolumn[
\icmltitle{Assessing the Robustness of Bayesian Dark Knowledge to Posterior Uncertainty}



\icmlsetsymbol{equal}{*}

\begin{icmlauthorlist}
\icmlauthor{Meet P. Vadera}{to}
\icmlauthor{Benjamin M. Marlin}{to}
\end{icmlauthorlist}

\icmlaffiliation{to}{College of Information and Computer Sciences, University of Massachusetts, Amherst}

\icmlcorrespondingauthor{Meet P. Vadera}{mvadera@cs.umass.edu}

\icmlkeywords{Machine Learning, ICML}

\vskip 0.3in
]



\printAffiliationsAndNotice{}  

\begin{abstract}
\emph{Bayesian Dark Knowledge} is a method for compressing the posterior predictive distribution of a neural network model into a more compact form. Specifically, the method attempts to compress a Monte Carlo approximation to the parameter posterior into a single network representing the posterior predictive distribution. Further, the authors show that this approach is successful in the classification setting using a student network whose architecture matches that of a single network in the teacher ensemble. In this work, we examine the robustness of Bayesian Dark Knowledge to higher levels of posterior uncertainty. We show that using a student network that matches the teacher architecture may fail to yield acceptable performance. We study an approach to close the resulting performance gap by increasing student model capacity.
\end{abstract}

\section{Introduction}

Deep learning models have shown promising results in the areas of computer vision, natural language processing, speech recognition, and more \cite{krizhevsky2012imagenet, graves2013hybrid, graves2013speech}. However, existing point estimation-based training methods for these models may result in predictive uncertainties that are not well calibrated, including the occurrence of confident errors. 

It is well-known that Bayesian inference can often provide more robust posterior predictive distributions in the classification setting compared to the use of point estimation-based training. However, the integrals required to perform Bayesian inference in neural network models are also well-known to be intractable. Monte Carlo methods provide one solution to representing neural network parameter posteriors as ensembles of networks, but this can require large amounts of both storage and compute time. 

To help overcome these problems, \citet{balan2015bayesian} introduced an interesting model training method referred to as \emph{Bayesian Dark Knowledge} that attempts to compress a traditional Monte Carlo approximation to the parameter posterior represented as an ensemble of teacher networks into a single student network representing the posterior predictive distribution. Further, the authors show that this approach is successful in the classification setting using a student network whose architecture matches that of a single network in the teacher ensemble. The Bayesian Dark Knowledge method also uses online learning of the student model based on single samples from the parameter posterior, resulting in a training algorithm that requires only twice as much space as a standard point estimate-based learning procedure. 

While the algorithmic properties and performance of the Bayesian Dark Knowledge method appear very promising in the classification setting based on the results presented by \citet{balan2015bayesian}, these results are limited to the MNIST data set in terms of real data. As can clearly be seen in the original paper, there is very little uncertainty in the posterior predictive distribution on MNIST using the network architecture selected by the authors. As a result, the performance of Bayesian Dark Knowledge in regimes of even moderate posterior predictive uncertainty has not been adequately examined. 

In this work, we examine the robustness of Bayesian Dark Knowledge to higher levels of posterior uncertainty and show that using a student network that matches the teacher architecture may result in a performance gap. Further, we show that the gap between the teacher ensemble and the student can be effectively reduced by increasing the capacity of the student. We examine two methods to  increase posterior predictive uncertainty for the MNIST data set, and present results on both fully connected and convolutional models.

\section{Related Work}
In cases where a Bayesian posterior distribution is not analytically tractable, approximations are required. The two most prevalent types of approximations studied in the machine learning literature are based on variational inference (VI) \cite{jordan1999introduction} and Markov Chain Monte Carlo (MCMC) methods \cite{Neal:1996:BLN:525544, welling2011bayesian}.

In VI, an auxiliary distribution $q_{\phi} (\theta)$  is defined to approximate the true parameter posterior $p(\theta | \pazocal{D})$. Next, we minimize the Kullback-Leibler (KL) divergence between $q_{\phi} (\theta)$ and $p(\theta | \pazocal{D})$ as a function of the variational parameters $\phi$. \citet{hinton1993keeping} presented the first study to apply VI to neural networks. However, the method presented scalability issues, and to tackle those issues, \citet{graves2011practical} presented an approach using stochastic VI. 

Another approach closely related to VI is expectation propagation (EP) \cite{minka2001expectation}. The main difference between VI and EP is that in VI, we minimize the $\textrm{KL}(q_{\phi} (\theta) || p(\theta | \pazocal{D}))$, while in EP we minimize $\textrm{KL}(p(\theta | \pazocal{D}) || q_{\phi} (\theta))$. \citet{soudry2014expectation} present an online EP algorithm for neural networks with the flexibility of representing both continuous and discrete weights. \citet{hernandez2015probabilistic}  present the probabilistic backpropagation (PBP) algorithm for approximate Bayesian learning of neural network models, which is an example of an assumed density filtering (ADF) algorithm that, like VI and EP, generally relies on simplified posterior densities.

To deal with the issues related to using simplified posterior densities (typically diagonal Gaussian) that arises in VB, EP, and PBP, \citet{louizos2017multiplicative} introduced the concept of \emph{Multiplicative Normalizing Flows}. Under Multiplicative Normalizing Flows, a chain of transformations are applied to parameters sampled from the approximate posterior $q_{\phi}(\theta)$ using some additional learnable parameters $\omega$. This technique helps to create samples imitating more complex distributions than standard mean-field inference. However, this requires the flow to be invertible.

A drawback of VB, EP, and ADF is that they all typically result in biased posterior estimates for complex posterior distributions. MCMC methods provide an alternative family of sampling-based posterior approximations that are unbiased, but are often computationally more expensive to use. Equation \ref{eq:posterior} shows the generic form of the parameter posterior under Bayesian inference where the denominator, referred to as the marginal likelihood or evidence, involves integrals with no closed-form solutions for neural network models.
\begin{align}
\allowdisplaybreaks[4]
\label{eq:posterior}
    p(\theta|\pazocal{D})& = \frac{p(\pazocal{D}|\theta)p(\theta)}{\int p(\pazocal{D}|\theta)p(\theta) d\theta}\\
\label{eq:posterior_predictive}
p(y| \mathbf{x}, \pazocal{D}) &= \int p(y|\mathbf{x}, \theta) p(\theta|\pazocal{D}) d\theta\\
\label{eq:MC_posterior_predictive}
&\approx \frac{1}{S}\sum_{t=1}^S p(y|\mathbf{x}, \theta_t); \;\;\;\;
\theta_t \sim p(\theta|\pazocal{D})
\end{align}
For prediction problems, the quantity of interest is not the parameter posterior, but the posterior predictive distribution, as shown in Equation \ref{eq:posterior_predictive}. This equation again involves intractable integrals, but can be approximated using a Monte Carlo average based on parameter values sampled from the posterior as seen in Equation  \ref{eq:MC_posterior_predictive}.

One of the earliest studies of Bayesian inference for neural network models  used Hamiltonian Monte Carlo (HMC) \cite{Neal:1996:BLN:525544} to provide a set of posterior samples. A bottleneck with this method is that it uses the full dataset when computing the gradient needed by HMC, which is problematic for large data sets. While this scalability problem has largely been solved by more recent methods such as stochastic gradient Langevin dynamics (SGLD) \cite{welling2011bayesian}, the problem of needing to retain and compute over a large set of samples when making predictions remains. 

Bayesian Dark Knowledge \cite{balan2015bayesian}, the focus of this work, is precisely aimed at reducing the computational complexity of making predictions using a sampling-based representation of the parameter posterior for neural networks. In particular, the method uses SGLD to approximate the posterior distribution using a set of posterior parameter samples. These samples can be thought of as an ensemble of neural network models with identical architectures, but different parameter values. This posterior ensemble is used as the teacher in a distillation process that trains a single student model to match the teacher ensemble's posterior predictive distribution.

Let $p(\theta| \lambda)$ be the prior distribution over the teacher neural network model parameters $\theta$. In this case, the prior distribution is a spherical Gaussian distribution centered at $0$ with precision denoted as $\lambda$. We define $\pazocal{S}$ to be a minibatch of size M drawn from $\pazocal{D}$. Let the total number of training samples in $\pazocal{D}$ be denoted by $N$. $\theta_t$ denotes the parameter set sampled for the teacher model at sampling iteration $t$, while $\eta_t$ denotes the learning rate for the teacher model at iteration $t$. The Langevin noise is denoted by $z_t \sim \pazocal{N}(0, \eta_t I )$.  Thus, the sampling update for SGLD can be written as:

{\footnotesize
\begin{equation}
\Delta \theta_{t+1} \!=\!\frac{\eta_{t}}{2}\!\left(\!\nabla_{\theta} \log p(\theta | \lambda)\!+\!\frac{N}{M} \sum_{i \in \pazocal{S}} \nabla_{\theta} \log p\left(y_{i} | x_{i}, \theta_{t}\right)\!\right)+ z_{t}.
\end{equation}
}

To train the student model, we generate another batch of samples $\pazocal{S'}$. $\pazocal{S'}$ is obtained by adding Gaussian noise of small magnitude to $\pazocal{S}$. The predictive distribution over $\{(\mathbf{x'}, y')\} \in \pazocal{S'}$ using the teacher model with parameters $\theta_{t+1}$ is given by $p(y'|\mathbf{x'}, \theta_{t+1})$. Similarly, the predictive distribution for a student model parameterized by $\omega_t$ is given as $p(y'|\mathbf{x'}, \omega_t)$. The objective for the student model is  $\pazocal{L}( \omega_t| \pazocal{S}', \theta_{t+1}) = \sum_{ \{(\mathbf{x'}, y')\} \in \pazocal{S'} }\textrm{KL}(p(y'|\mathbf{x'}, \omega_t) || p(y'|\mathbf{x'}, \theta_{t+1}))$ and we run a single optimization iteration on it to obtain $\omega_{t+1}$. This process of sequentially computing $\theta_{t+1}$ and $\omega_{t+1}$ is repeated until convergence. 

Finally, we note that with the advent of \emph{Generative Adversarial Networks} \cite{goodfellow2014generative}, there has also been work on generative models for approximating posterior sampling. \citet{wang2018adversarial}  and \citet{henning2018approximating} both propose methods for learning to generate samples that mimic those produced by SGLD. However, while these approaches may provide a speed-up relative to running SGLD itself, the resulting samples must still be used in a Monte Carlo average to to compute a posterior predictive distribution in the case of Bayesian neural networks. This is again a potentially costly operation and is exactly the computation that  Bayesian Dark Knowledge addresses.

\section{Assessing  Robustness to Uncertainty}
As noted in the introduction, the original empirical investigation of Bayesian Dark Knowledge for classification focused on the MNIST data set. However, the models fit to the MNIST data set have very low posterior uncertainty and we argue that it is thus a poor benchmark for assessing the performance of Bayesian Dark Knowledge. In this section, we investigate two orthogonal modifications of the standard MNIST data  set: reducing the training set size and masking regions of the input images. Our goal is to produce a range of benchmark problems with varying posterior predictive uncertainty. 

\textbf{Data Set Manipulations:} The full MNIST dataset consists of 60,000 training images and 10,000 test images, each of size $28 \times 28$, distributed among 10 classes \cite{lecun1998mnist}. As a first manipulation, we consider sub-sampling the labeled training data to include 10,000, 20,000, 30,000 or all 60,000 data cases when performing posterior sampling for the teacher model. Importantly, we use all 60,000 unlabeled training cases in the distillation process, which does not require labeled instances. This allows us de-couple the impact of reduced labeled training data on posterior predictive distributions from the effect of the amount of unlabeled data available for distillation.

\begin{figure}[!htbp]
    \centering
    \subfigure[Original images]{\includegraphics[width=0.18\textwidth]{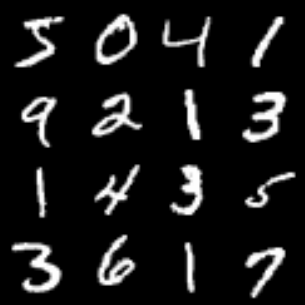}}
    \hspace{1cm}
    \subfigure[Processed images]{\includegraphics[width=0.18\textwidth]{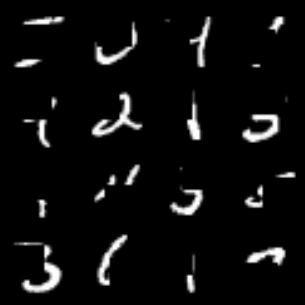}}
    \caption{Inputs after data pre-processing step with $m=14$.\vspace{-1em}}
    \label{fig:mnist}
\end{figure}

As a second manipulation, we generate images with occlusions by randomly masking out parts of each available training and test image. For generating such images, we randomly choose a square $m\times m$ region (mask) and set the value for pixels in that region to 0. Thus, the masking rate for a $28 \times 28$ MNIST image corresponding to the mask of size $m\times m$ is given by $r = \frac{m \times m}{28 \times 28}$. We illustrate original and masked data in Figure \ref{fig:mnist}. We consider a range of square masks resulting in masking rates between 0\% and 86.2\%.

\textbf{Models:} We use a fully-connected neural network of architecture $784-400-400-10$ for both teacher and student with $ReLU$ non-linearities at the hidden layers. The only addition we make to the student model is using dropout layers for regularization. We run the distillation procedure using the following hyperparameters: fixed teacher learning rate $\eta_t = 4 \times 10^{-6}$, teacher prior precision $\lambda = 10$, initial student learning rate $\rho_t = 10^{-3}$, student dropout rate $p=0.5$, burn-in iterations $B=1000$, thinning interval $\tau = 100$, and total training iterations $T= 10^6$. For training the student model, we use the \emph{Adam} algorithm and set a learning schedule for the student such that it halves its learning rate every 100 epochs. For computing the teacher model's performance on the test set, we use all samples after burn-in  and average the softmax outputs generated by each sample. For the student model, we use the final model obtained after training. We consider the negative log-likelihood (NLL) on the test set as a metric for model performance (zero is the minimum possible value). 

\begin{figure*}[!htbp]
    \centering
    \subfigure{\includegraphics[width=0.4\textwidth]{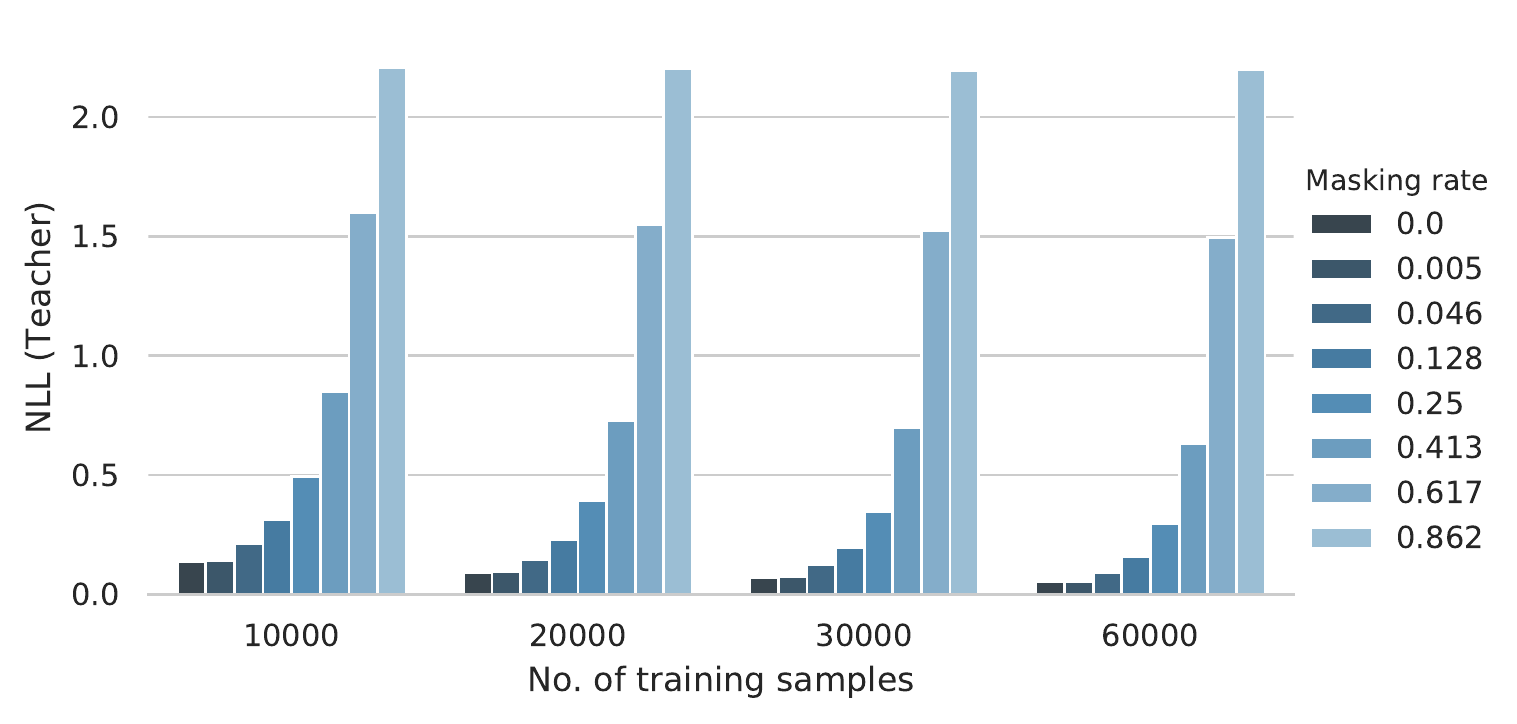}}
    \subfigure{\includegraphics[width=0.4\textwidth]{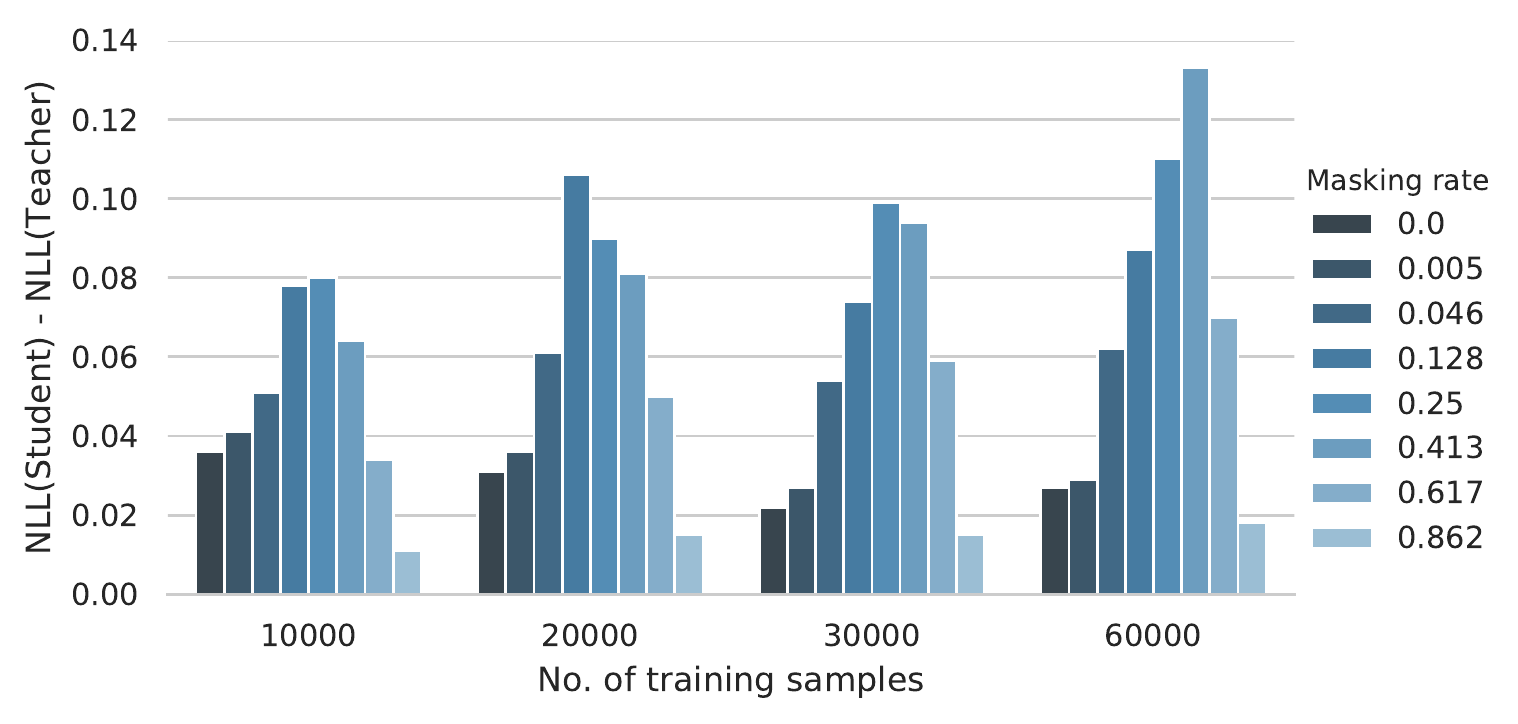}}
    \caption{Left: NLL on test set using teacher model. Right: Difference in NLL on test set between student and teacher.}
    \label{fig:delta_plot}
\end{figure*}

\textbf{Results:} The NLL of the teacher for different labeled training data set sizes and masking rates is shown in Figure \ref{fig:delta_plot}(left) along with the difference between the NLL using the student and teacher in Figure \ref{fig:delta_plot}(right).
As we can see, the teacher obtains very low NLL values when the data set is at full size and no masking is applied, which replicates the results shown by \citet{balan2015bayesian}. As the amount of labeled training data decreases, the NLL generally increases, but only slowly. On the other hand, as the masking rate increases, the NLL of the teacher increases to nearly the maximum possible value, indicating that the posterior predictive distribution is approximately uniform. 

Next,  we consider the gap between the teacher ensemble and the student model. First, we see that the gap is very low for the case of no  masking and 60,000 training instances, again replicating the results of \citet{balan2015bayesian}. We can also see that the gaps between the teacher and the student become much larger for intermediate masking rates. However, we note that the gap in performance between the student and the teacher is not monotonically increasing with the masking rate. The reason for this is that as the masking rate goes to one, we can see from Figure 2(left) that the the posterior predictive distribution becomes very close to uniform. It is in fact very easy for the student to learn to output the uniform distribution for every input  image, so the gap between the teacher and the student becomes minimal. 

The primary conclusion of this experiment is that compression of the full posterior predictive distribution into a student model of the same capacity as the teacher network is failing under some experimental conditions. Specifically, the original Bayesian Dark Knowledge approach appears  to not be robust to the impact that the masking operation has on posterior predictive uncertainty. We perform additional experiments on convolutional models and present a similar set of results in Appendix A. Appendix A also includes posterior predictive entropy plots verifying the effect of masking on posterior uncertainty.

\section{Improving Bayesian Dark Knowledge}
In this section, we show that increasing student model capacity can  close some of the performance gap we highlighted in the previous section. We investigate both fully-connected and convolutional models.

\textbf{Fully-Connected Neural Networks:} We first define a base architecture that we leverage to build larger models. The base architecture is a $784-400-400-1$ model, and is also used for the teacher model.
Next, we use a multiplication factor $K$ to multiply the number of units in the hidden layers to generate models of different capacities.  For a given value of $K$, the student model architecture is given by: $784-\floor*{400 \cdot K}-\floor*{400 \cdot K}-1$. We choose the experimental setting of $N=60,000$ and masking rate of $0.25$, as this resulted in a large performance gap in the previous  section. We use the same hyperparameters as stated in the previous section and examine the the performance gaps in distillation as we vary $K$. 

\begin{figure}[t]
    \centering
    \includegraphics[width=0.48\textwidth]{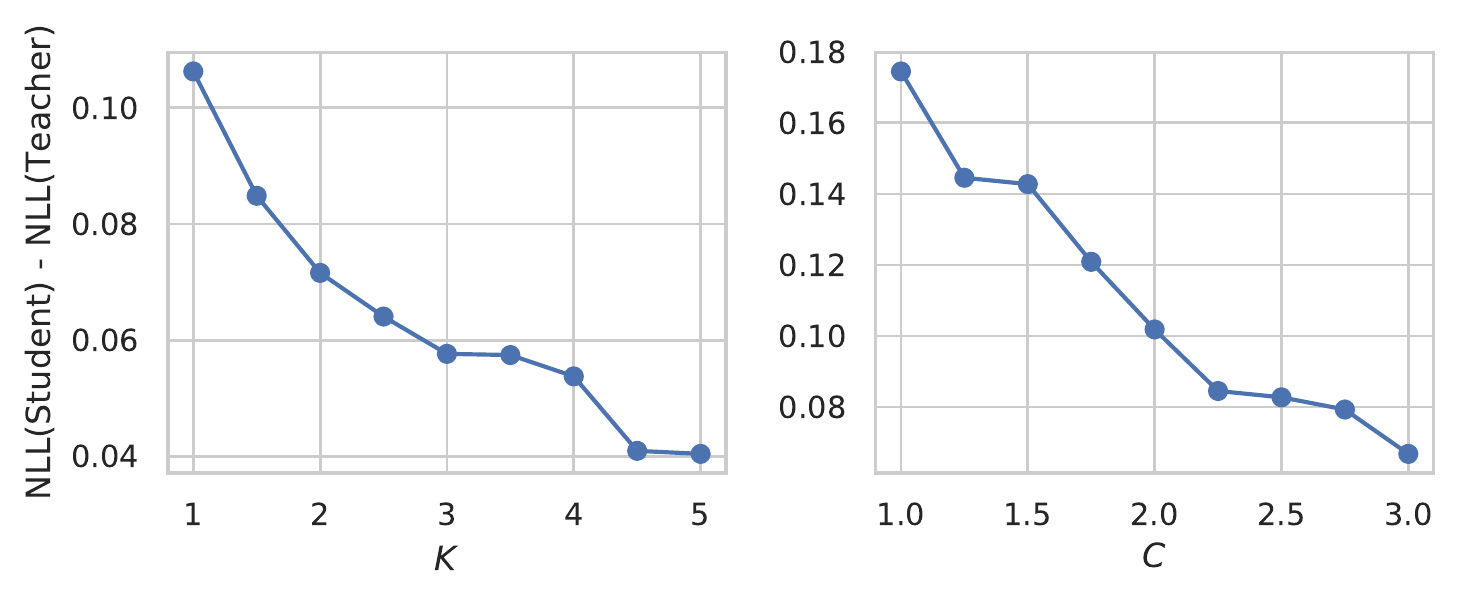}
    \vspace{-.2in}
    \caption{Comparison of the gap in performance between the student and the teacher for different student model sizes. The left plot shows results for the FCNN. The right plot shows results for CNN.\vspace{-1.5em}}
    \label{fig:model_expansion_results}
\end{figure}

The results for this experiment are shown in Figure \ref{fig:model_expansion_results}(left).
As we can see, the gap between the NLL of student and teacher drops by about $60\%$ as we increase the value of $K$ to 5. This shows that increasing the model capacity helps improve the performance of the student model significantly.

\textbf{Convolutional Neural Networks:}
For CNNs, we follow the same strategy as we did in the case of FCNNs. We define a base architecture and then build on the existing base architecture by increasing the number of filters and hidden units. The base architecture for our CNN model is: Conv(num\_kernels = 10, kernel\_size = 4, stride = 1) - MaxPool(kernel\_size=2) - Conv(num\_kernels = 20, kernel\_size = 4, stride = 1) - MaxPool(kernel\_size=2) - FC (80) - FC (10), and is also used for the teacher model. For increasing the model capacity of student models, we use a multiplication factor of $C$ to multiply the number of kernels in convolutional layers and the number of hidden units for fully connected layer in the base architecture. The new architecture is given as: Conv(num\_kernels = $\floor*{10 \cdot C}$, kernel\_size = 4, stride = 1) - MaxPool(kernel\_size=2) - Conv(num\_kernels = $\floor*{20 \cdot C}$ = , kernel\_size = 4, stride = 1) - MaxPool(kernel\_size=2) - FC ($\floor*{80 \cdot C}$) - FC (10). We again perform our experiment with the setting $N=60,000$ and masking rate of $0.25$ for different values of $C$. The results are shown in the Figure \ref{fig:model_expansion_results}(right).

The gap between the NLL of student and teacher reduces by approximately $60\%$ as we set $C = 3$. This is again a significant improvement in the performance of the student model as compared to the base case where $C = 1$.  

\section{Conclusions \& Future Directions}
Our results show that the performance of the original Bayesian Dark Knowledge method can degrade under conditions with moderate posterior uncertainty, and that the gap in performance between the student and teacher can  be attributed to insufficient capacity in the student model. There are  many potential directions for future work given that the architecture of the  student model may require customization  relative  to the teacher to provide sufficient capacity while controlling for computation and storage costs.

\section*{Acknowledgements}

The research reported in this article was sponsored in part by the Army Research Laboratory under Cooperative Agreement W911NF-17-2-0196. The views and conclusions contained in this document are those of the authors and should not be interpreted as representing the official policies, either expressed or implied, of the Army Research Laboratory or the US government. The US government is authorized to reproduce and distribute reprints for government purposes notwithstanding any copyright notation here on.

\bibliography{references}
\bibliographystyle{icml2019}
\begin{table*}[t!]
\caption{Distillation performance with different data sizes ($N$), mask sizes ($m$) and masking rates ($r$). $\Delta$ denotes the performance gap between teacher and student on test set. Note that the for every row, the test set also has the same masking rate. The left table shows results for fully connected networks. The right table shows results for convolutional networks.}
\vspace{0.15in}
\begin{minipage}{0.45\hsize}\centering
\begin{tabular}{cccccc}
\toprule
$N$ & $m$ & $r$ & \begin{tabular}[c]{@{}c@{}}NLL\\ (Teacher)\end{tabular} & \begin{tabular}[c]{@{}c@{}}NLL\\ (Student)\end{tabular} & $\Delta$ \\ \toprule
\multirow{8}{*}{10000} & 0 & 0.000 & 0.138 & 0.173 & 0.036 \\ 
 & 2 & 0.005 & 0.140 & 0.181 & 0.041 \\ 
 & 6 & 0.046 & 0.213 & 0.264 & 0.051 \\ 
 & 10 & 0.128 & 0.314 & 0.392 & 0.078 \\ 
 & 14 & 0.250 & 0.493 & 0.573 & 0.080 \\ 
 & 18 & 0.413 & 0.851 & 0.915 & 0.064 \\ 
 & 22 & 0.617 & 1.600 & 1.634 & 0.034 \\ 
 & 26 & 0.862 & 2.205 & 2.216 & 0.011 \\ \midrule
\multirow{8}{*}{20000} & 0 & 0.000 & 0.090 & 0.121 & 0.031 \\ 
 & 2 & 0.005 & 0.094 & 0.130 & 0.036 \\ 
 & 6 & 0.046 & 0.147 & 0.209 & 0.061 \\ 
 & 10 & 0.128 & 0.228 & 0.334 & 0.106 \\ 
 & 14 & 0.250 & 0.392 & 0.482 & 0.090 \\ 
 & 18 & 0.413 & 0.729 & 0.810 & 0.081 \\ 
 & 22 & 0.617 & 1.550 & 1.600 & 0.050 \\ 
 & 26 & 0.862 & 2.204 & 2.218 & 0.015 \\ \midrule
\multirow{8}{*}{30000} & 0 & 0.000 & 0.071 & 0.093 & 0.022 \\ 
 & 2 & 0.005 & 0.073 & 0.099 & 0.027 \\ 
 & 6 & 0.046 & 0.123 & 0.177 & 0.054 \\ 
 & 10 & 0.128 & 0.195 & 0.269 & 0.074 \\ 
 & 14 & 0.250 & 0.347 & 0.446 & 0.099 \\ 
 & 18 & 0.413 & 0.697 & 0.791 & 0.094 \\ 
 & 22 & 0.617 & 1.524 & 1.584 & 0.059 \\ 
 & 26 & 0.862 & 2.194 & 2.209 & 0.015 \\ \midrule
\multirow{8}{*}{60000} & 0 & 0.000 & 0.052 & 0.080 & 0.027 \\ 
 & 2 & 0.005 & 0.055 & 0.084 & 0.029 \\ 
 & 6 & 0.046 & 0.092 & 0.155 & 0.062 \\ 
 & 10 & 0.128 & 0.159 & 0.246 & 0.087 \\ 
 & 14 & 0.250 & 0.296 & 0.407 & 0.110 \\ 
 & 18 & 0.413 & 0.630 & 0.763 & 0.133 \\ 
 & 22 & 0.617 & 1.495 & 1.565 & 0.070 \\ 
 & 26 & 0.862 & 2.197 & 2.215 & 0.018 \\ \bottomrule
\end{tabular}
\label{tab:masked_mnist_results}
\end{minipage}\hfill
\begin{minipage}{0.45\hsize}\centering
\begin{tabular}{cccccc}
\toprule
$N$ & $m$ & $r$ & \begin{tabular}[c]{@{}c@{}}NLL\\ (Teacher)\end{tabular} & \begin{tabular}[c]{@{}c@{}}NLL\\ (Student)\end{tabular} & $\Delta$ \\ \toprule
\multirow{8}{*}{10000} & 0 & 0.000 & 0.138 & 0.173 & 0.036 \\ 
 & 2 & 0.005 & 0.140 & 0.181 & 0.041 \\ 
 & 6 & 0.046 & 0.213 & 0.264 & 0.051 \\ 
 & 10 & 0.128 & 0.314 & 0.392 & 0.078 \\ 
 & 14 & 0.250 & 0.493 & 0.573 & 0.080 \\ 
 & 18 & 0.413 & 0.851 & 0.915 & 0.064 \\ 
 & 22 & 0.617 & 1.600 & 1.634 & 0.034 \\ 
 & 26 & 0.862 & 2.205 & 2.216 & 0.011 \\ \midrule
\multirow{8}{*}{20000} & 0 & 0.000 & 0.090 & 0.121 & 0.031 \\ 
 & 2 & 0.005 & 0.094 & 0.130 & 0.036 \\ 
 & 6 & 0.046 & 0.147 & 0.209 & 0.061 \\ 
 & 10 & 0.128 & 0.228 & 0.334 & 0.106 \\ 
 & 14 & 0.250 & 0.392 & 0.482 & 0.090 \\ 
 & 18 & 0.413 & 0.729 & 0.810 & 0.081 \\ 
 & 22 & 0.617 & 1.550 & 1.600 & 0.050 \\ 
 & 26 & 0.862 & 2.204 & 2.218 & 0.015 \\ \midrule
\multirow{8}{*}{30000} & 0 & 0.000 & 0.071 & 0.093 & 0.022 \\ 
 & 2 & 0.005 & 0.073 & 0.099 & 0.027 \\ 
 & 6 & 0.046 & 0.123 & 0.177 & 0.054 \\ 
 & 10 & 0.128 & 0.195 & 0.269 & 0.074 \\ 
 & 14 & 0.250 & 0.347 & 0.446 & 0.099 \\ 
 & 18 & 0.413 & 0.697 & 0.791 & 0.094 \\ 
 & 22 & 0.617 & 1.524 & 1.584 & 0.059 \\ 
 & 26 & 0.862 & 2.194 & 2.209 & 0.015 \\ \midrule
\multirow{8}{*}{60000} & 0 & 0.000 & 0.052 & 0.080 & 0.027 \\ 
 & 2 & 0.005 & 0.055 & 0.084 & 0.029 \\ 
 & 6 & 0.046 & 0.092 & 0.155 & 0.062 \\ 
 & 10 & 0.128 & 0.159 & 0.246 & 0.087 \\ 
 & 14 & 0.250 & 0.296 & 0.407 & 0.110 \\ 
 & 18 & 0.413 & 0.630 & 0.763 & 0.133 \\ 
 & 22 & 0.617 & 1.495 & 1.565 & 0.070 \\ 
 & 26 & 0.862 & 2.197 & 2.215 & 0.018 \\ \bottomrule
\end{tabular}
\label{tab:cnn-masking_results}

\end{minipage}
\end{table*}

\appendix

\section{Additional results on Masked MNIST Data}
\begin{figure*}[t]
    \centering
    \subfigure{\includegraphics[width=0.45\textwidth]{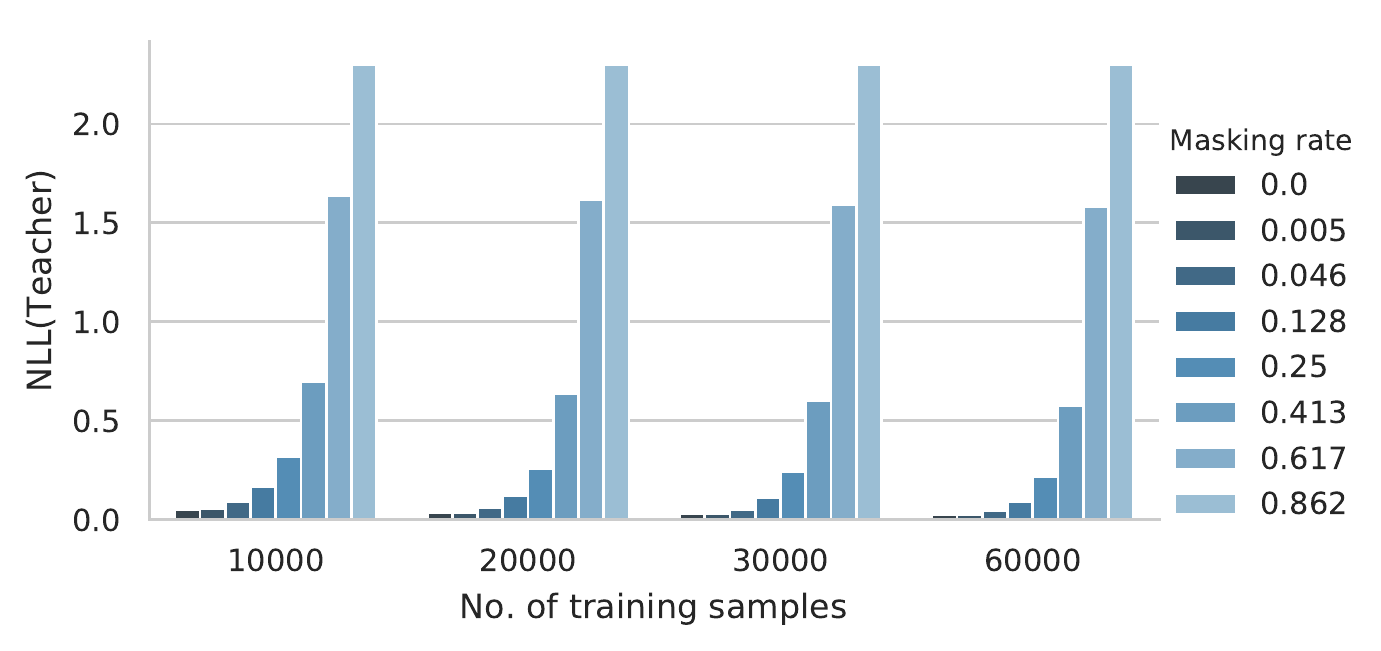}}
    \subfigure{\includegraphics[width=0.45\textwidth]{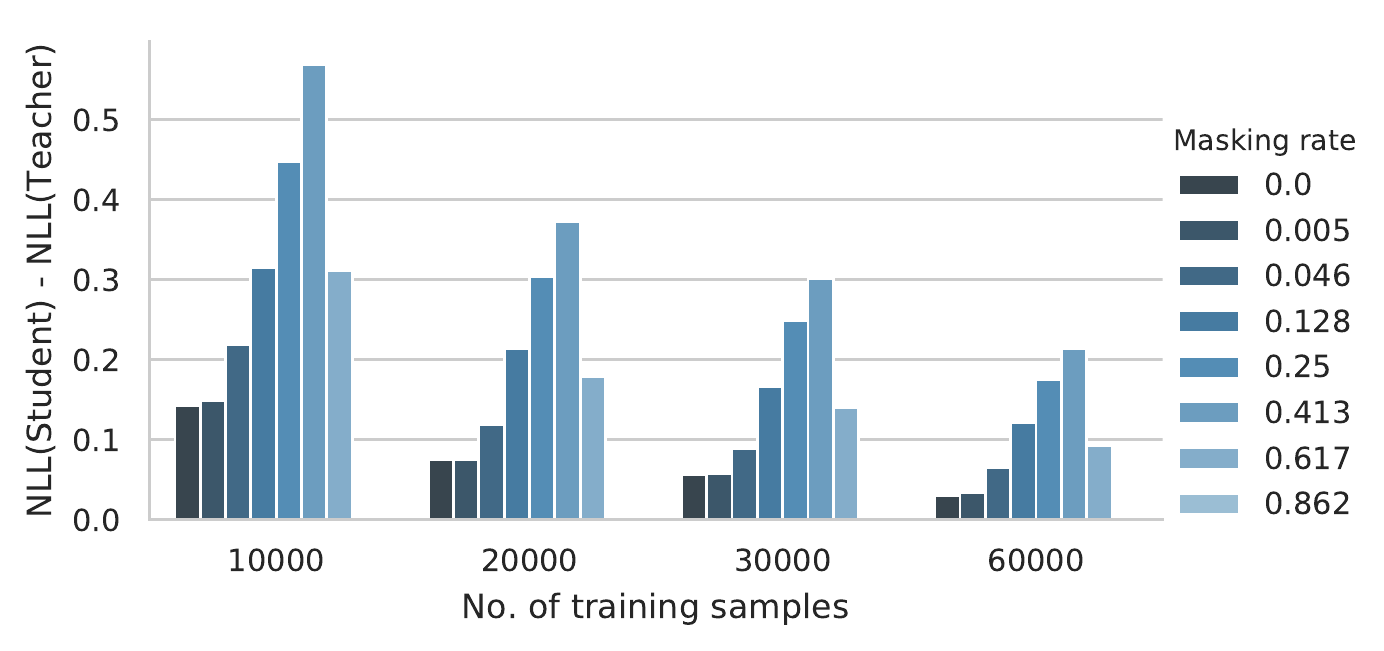}}
    \caption{Left: NLL on the test set using a CNN teacher model. Right: Difference in NLL on the test set between student CNN and teacher CNN.}
    \label{fig:cnn_delta_plot}

    \centering
    \subfigure[$N=10000$]{\includegraphics[width=0.32\textwidth]{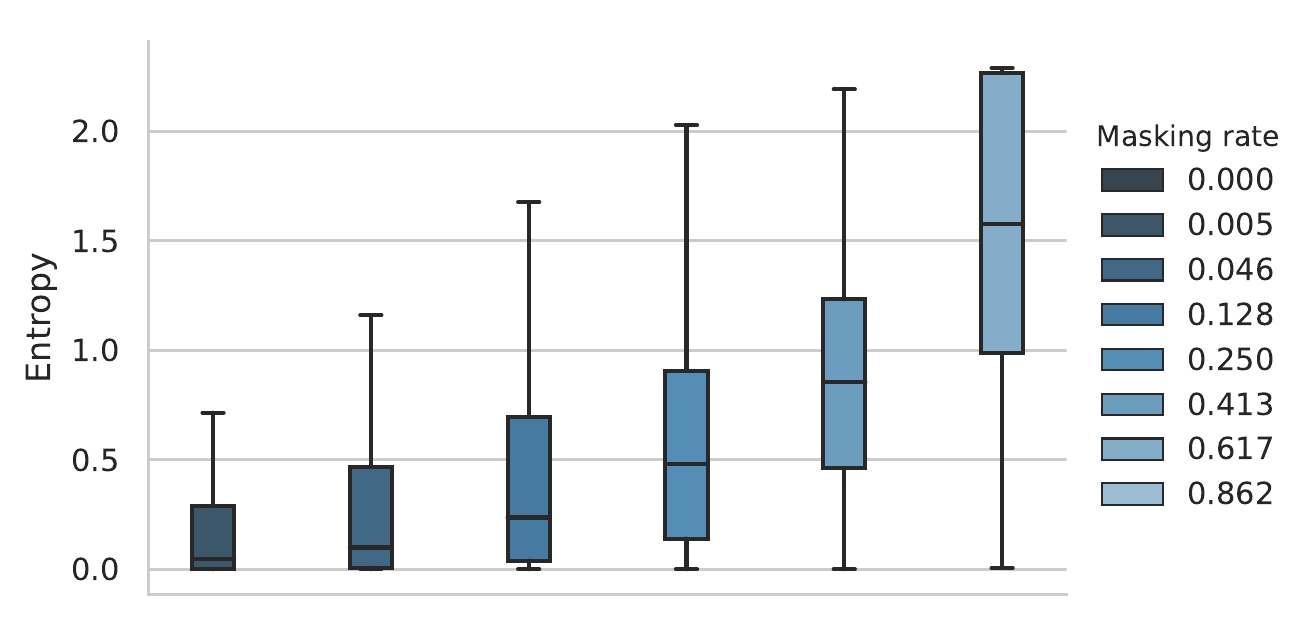}}
    \subfigure[$N=20000$]{\includegraphics[width=0.32\textwidth]{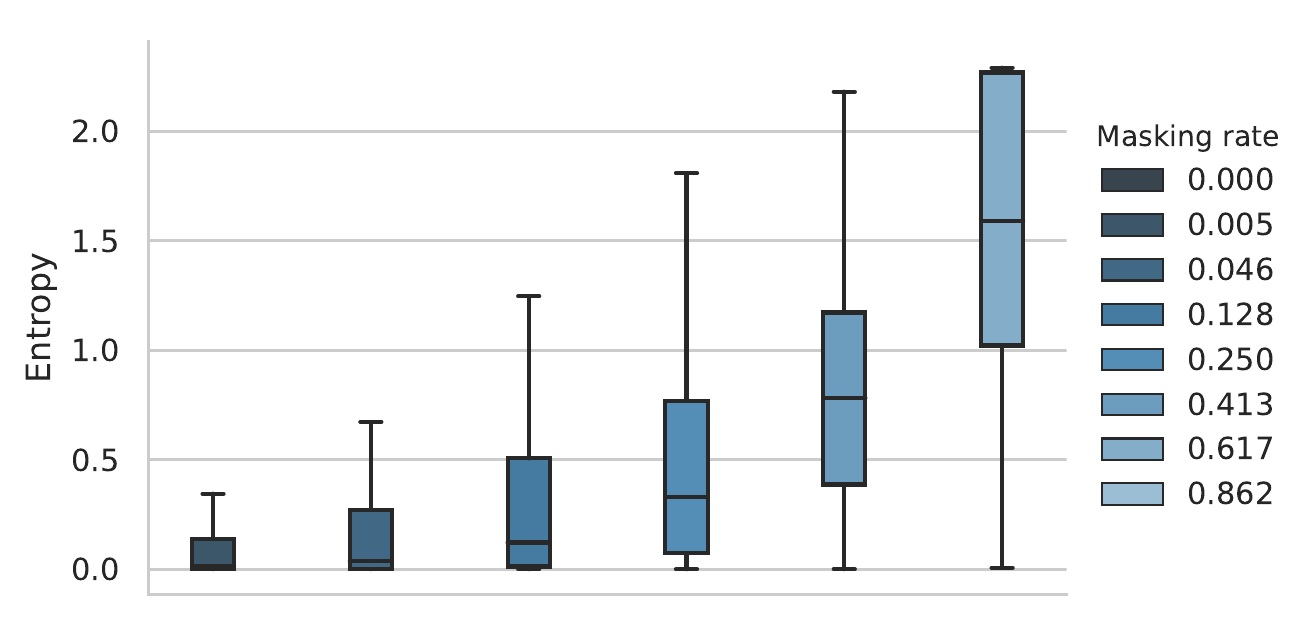}}
    \subfigure[$N=60000$]{\includegraphics[width=0.32\textwidth]{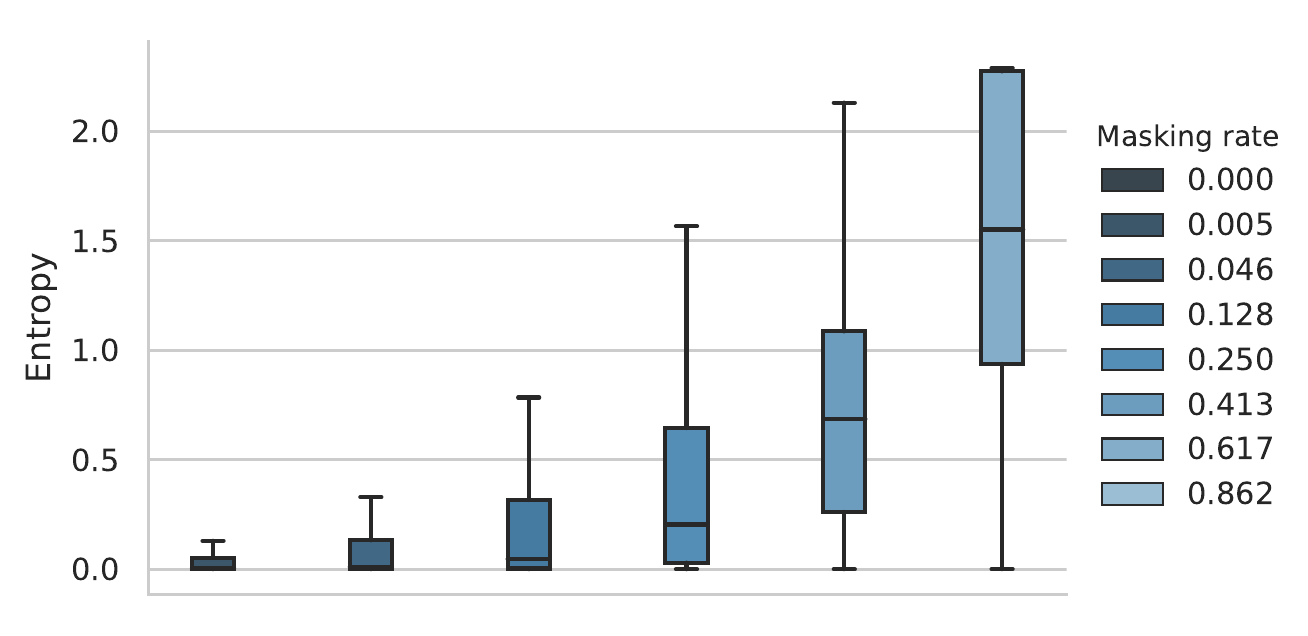}}
    \caption{Entropy of the posterior predictive distribution of the teacher model on the test set using \textbf{fully-connected neural networks.}}
    \label{fig:boxplot_fcnn}

    \centering
    \subfigure[$N=10000$]{\includegraphics[width=0.32\textwidth]{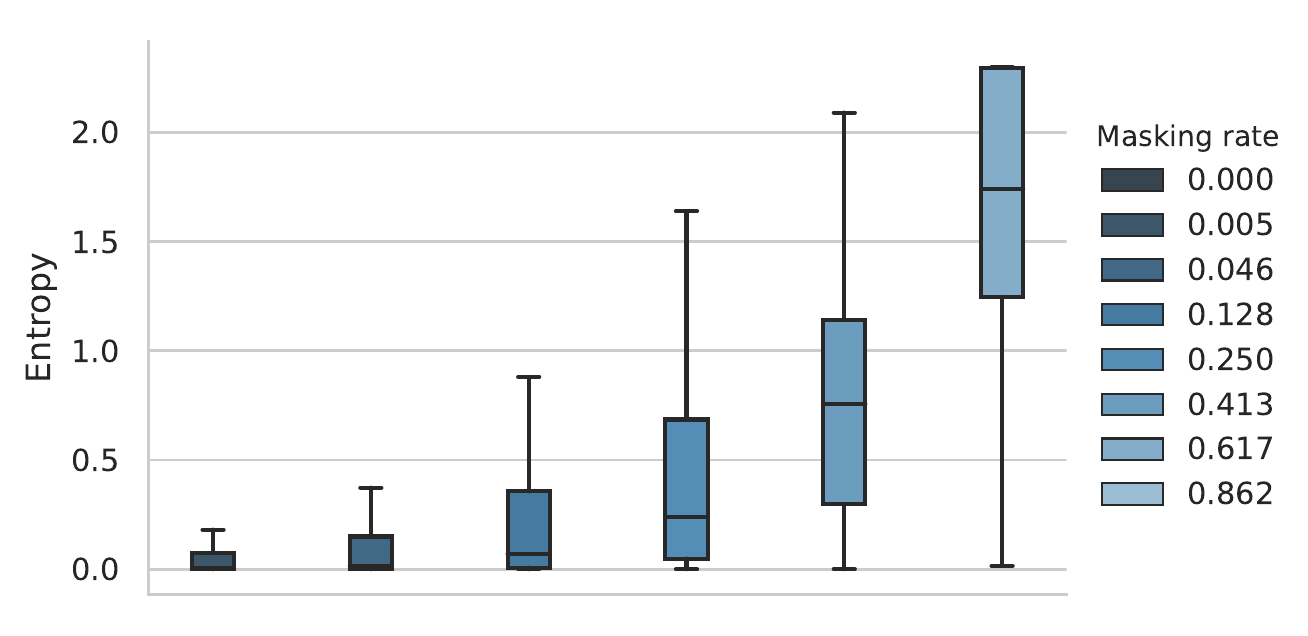}}
    \subfigure[$N=20000$]{\includegraphics[width=0.32\textwidth]{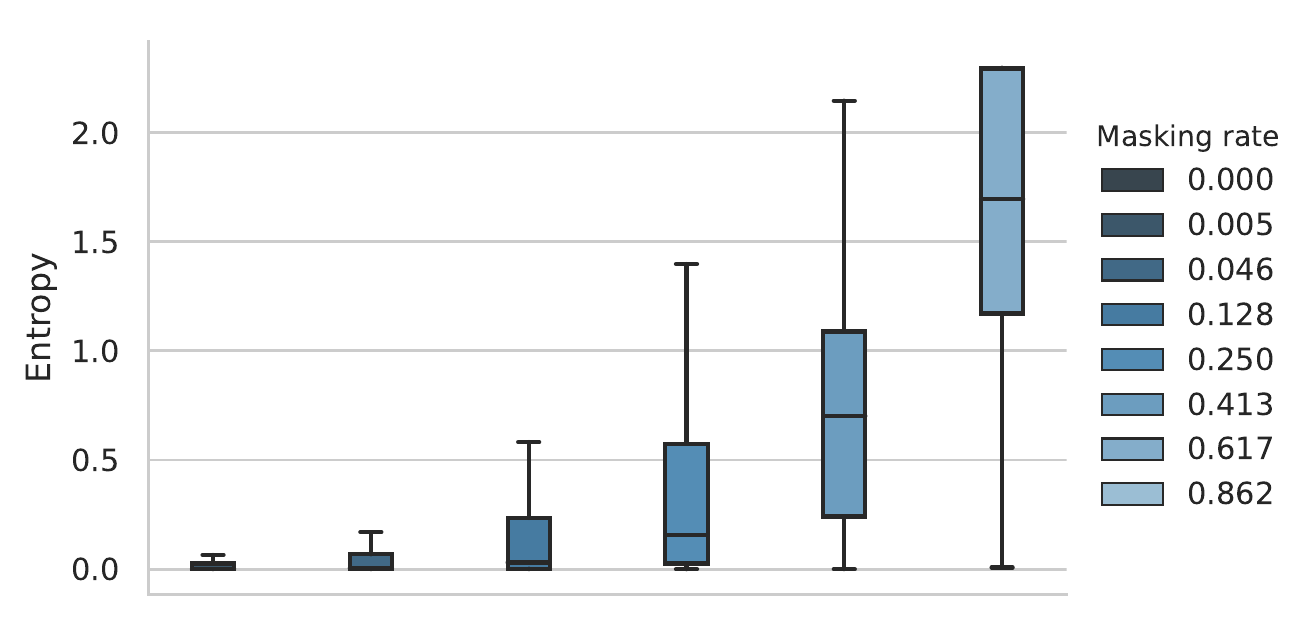}}
    \subfigure[$N=60000$]{\includegraphics[width=0.32\textwidth]{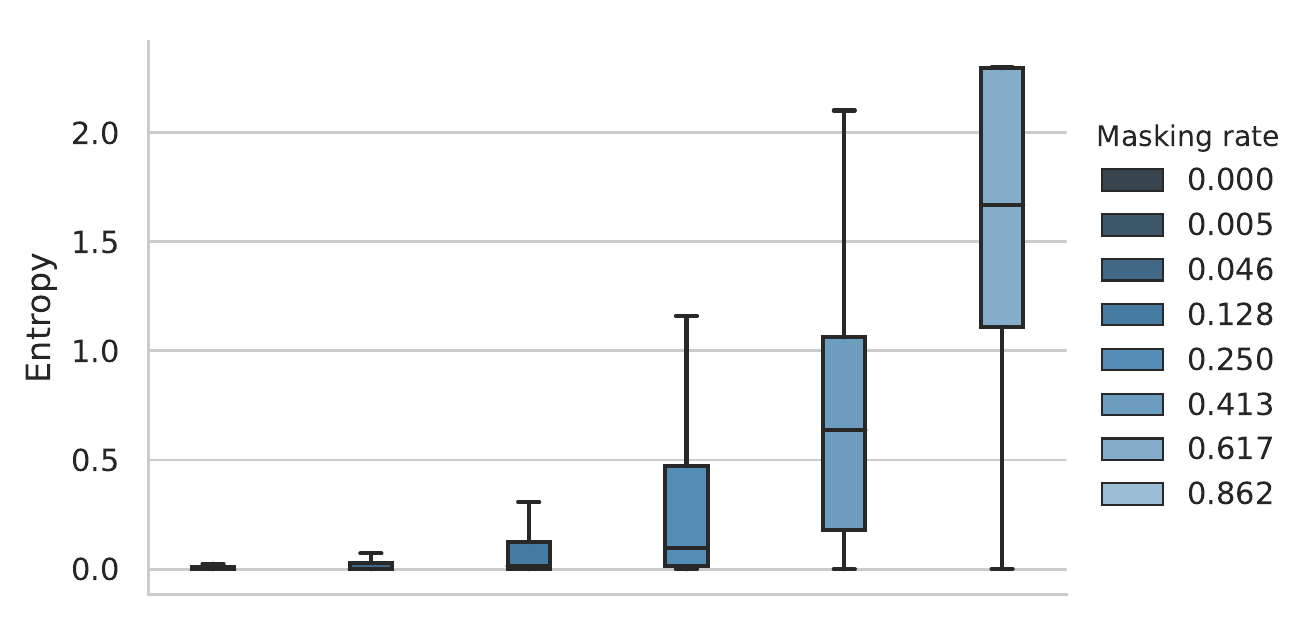}}
    \caption{Entropy of the posterior predictive distribution of the teacher model on the test set using \textbf{convolutional neural networks.}}
    \label{fig:boxplot_cnn}
\end{figure*}

Similar to Figure \ref{fig:delta_plot}, we show the trends in distillation performance for convolutional neural networks in Figure \ref{fig:cnn_delta_plot}. The detailed results from the experiments performed using both types of models are given in Table \ref{tab:masked_mnist_results}. Note that the experiments to evaluate the robustness of Bayesian Dark Knowledge for CNNs are performed using the following CNN architecture for both the teacher and the student model: Conv(num\_kernels = 10, kernel\_size = 4, stride = 1) - MaxPool(kernel\_size=2) - Conv(num\_kernels = 20, kernel\_size = 4, stride = 1) - MaxPool(kernel\_size=2) - FC (80) - FC (10). This has also been chosen as the base architecture for experiments performed in Section 4. The hyperparameters are the same as used in Section 3 for fully-connected networks. We observe that the performance trends for distillation as we vary number of training samples and masking rate are very similar for both fully-connected networks and convolutional neural networks. 
To probe deeper into the posterior uncertainty, we present box plots of the posterior predictive distribution on the test set for the teacher model in Figure \ref{fig:boxplot_fcnn}, and Figure \ref{fig:boxplot_cnn} for fully-connected and convolutional  networks respectively. We see that masking has a significant impact on the posterior uncertainty. The entropy results, when complemented with the results in Table \ref{tab:masked_mnist_results} demonstrate that the original Bayesian Dark Knowledge approach is not robust to posterior uncertainty.

\end{document}